\documentclass[11pt,a4paper]{article}
\usepackage[hyperref]{acl2019}
\usepackage{times}
\usepackage{latexsym}

\usepackage{helvet}  
\usepackage{courier}  
\usepackage{graphicx}  
\usepackage{makecell}
\usepackage{multirow}
\usepackage{algorithm}
\usepackage{algorithmic}
\usepackage{amsmath}
\usepackage{colortbl}
\usepackage{amsfonts}

\usepackage{url}

\aclfinalcopy 

\title{Towards Explainable NLP: A Generative Explanation \\Framework for Text Classification}
\author{Hui Liu$^1$, Qingyu Yin$^2$, William Yang Wang$^3$ \\
 $^1$ Peking University, China \\
  $^2$ Harbin Institute of Technology, China \\
  $^3$ University of California, Santa Barbara, USA\\
  {\tt layneliuhui@gmail.com } \\
  {\tt qyyin@ir.hit.edu.cn} \\
  {\tt william@cs.ucsb.edu} \\
  }

\date{}

\begin{document}
\maketitle
\begin{abstract}
Building explainable systems is a critical problem in the field of Natural Language Processing (NLP), since most machine learning models provide no explanations for the predictions. Existing approaches for explainable machine learning systems tend to focus on interpreting the outputs or the connections between inputs and outputs. However, the \emph{fine-grained information} (e.g. textual explanations for the labels) is often ignored, and the systems do not explicitly generate the human-readable explanations. To solve this problem, we propose a novel generative explanation framework that learns to make classification decisions and generate fine-grained explanations at the same time. More specifically, we introduce the explainable factor and the minimum risk training approach that learn to generate more reasonable explanations. We construct two new datasets that contain summaries, rating scores, and fine-grained reasons. We conduct experiments on both datasets, comparing with several strong neural network baseline systems. Experimental results show that our method surpasses all baselines on both datasets, and is able to generate concise explanations at the same time.
\end{abstract}

\section{Introduction}

Deep learning methods have produced state-of-the-art results in many natural language processing (NLP) tasks \cite{vaswani2017attention,yin2018deep,Peters:2018,wang2018multi,P18-1175,P18-1130}. Though these deep neural network models achieve impressive performance, it is relatively difficult to convince people to trust the predictions of such neural networks since they are actually black boxes for human beings \cite{samek2017explainable}. For instance, if an essay scoring system only tells the scores of a given essay without providing explicit reasons, the users can hardly be convinced of the judgment. Therefore, the ability to explain the rationale is essential for a NLP system, a need which requires traditional NLP models to provide human-readable explanations.

In recent years, lots of works have been done to solve text classification problems, but just a few of them have explored the explainability of their systems \cite{NIPS2018_8163,ouyang2018improving}. \newcite{ribeiro2016should} try to identify an interpretable model over the interpretable representation that is locally faithful to the classifier. \newcite{samek2017explainable} use heatmap to visualize how much each hidden element contributes to the predicted results. Although these systems are somewhat promising, they typically do not consider fine-grained information that may contain information for interpreting the behavior of models. However, if a human being wants to rate a product, s/he may first write down some reviews, and then score or summarize some attributes of the product, like price, packaging, and quality. Finally, the overall rating for the product will be given based on the fine-grained information. Therefore, it is crucial to build trustworthy explainable text classification models that are capable of explicitly generating fine-grained information for explaining their predictions.

To achieve these goals, in this paper, we propose a novel generative explanation framework for text classification, where our model is capable of not only providing the classification predictions but also generating fine-grained information as explanations for decisions. The novel idea behind our hybrid generative-discriminative method is to explicitly capture the fine-grained information inferred from raw texts, utilizing the information to help interpret the predicted classification results and improve the overall performance.
Specifically, we introduce the notion of an explainable factor and a minimum risk training method that learn to generate reasonable explanations for the overall predict results. Meanwhile, such a strategy brings strong connections between the explanations and predictions, which in return leads to better performance. To the best of our knowledge, we are the first to explicitly explain the predicted results by utilizing the abstractive generative fine-grained information.

In this work, we regard the summaries (texts) and rating scores (numbers) as the fine-grained information. Two datasets that contain these kinds of fine-grained information are collected to evaluate our method. More specifically, we construct a dataset crawled from a website called PCMag\footnote{\url{https://www.pcmag.com/}}. Each item in this dataset consists of three parts: a long review text for one product, three short text comments (respectively explains the property of the product from positive, negative and neutral perspectives) and an overall rating score. We regard the three short comments as fine-grained information for the long review text. Besides, we also conduct experiments on the Skytrax User Reviews Dataset\footnote{\url{https://github.com/quankiquanki/skytrax-reviews-dataset}}, where each case consists of three parts: a review text for a flight, five sub-field rating scores (seat comfortability, cabin stuff, food, in-flight environment, ticket value) and an overall rating score. As for this dataset, we regard the five sub-field rating scores as fine-grained information for the flight review text.

Empirically, we evaluate our model-agnostic method on several neural network baseline models \cite{kim2014convolutional,liu2016recurrent,acl2018zhou} for both datasets. Experimental results suggest that our approach substantially improves the performance over baseline systems, illustrating the advantage of utilizing fine-grained information. Meanwhile, by providing the fine-grained information as explanations for the classification results, our model is an understandable system that is worth trusting. Our major contributions are three-fold:

\begin{itemize}
	\item We are the first to leverage the generated fine-grained information for building a generative explanation framework for text classification, propose an explanation factor, and introduce minimum risk training for this hybrid generative-discriminative framework;
	
	\item We evaluate our model-agnostic explanation framework with different neural network architectures, and show considerable improvements over baseline systems on two datasets;
	
	\item We provide two new publicly available explainable NLP datasets that contain fine-grained information as explanations for text classification.
	
\end{itemize}


\section{Task Definition and Notations}
The research problem investigated in this paper is defined as: How can we generate fine-grained explanations for the decisions our classification model makes? To answer this question, we may first investigate what are good fine-grained explanations. For example, in sentiment analysis, if a product $A$ has three attributes: i.e., quality, practicality, and price. Each attribute can be described as ``HIGH'' or ``LOW''. And we want to know whether $A$ is a ``GOOD'' or ``BAD'' product. If our model categorizes $A$ as ``GOOD'' and it tells that the quality of $A$ is ``HIGH'', the practicality is ``HIGH'' and the price is ``LOW'', we can regard these values of attributes as good explanations that illustrate why the model judges $A$ to be ``GOOD''. On the contrary, if our model produces the same values for the attributes, but it tells that $A$ is a ``BAD'' product, we then think the model gives bad explanations. Therefore, for a given classification prediction made by the model, we would like to explore more on the fine-grained information that can explain why it comes to such a decision for the current example. Meanwhile, we also want to figure out whether the fine-grained information inferred from the input texts can help improve the overall classification performance.

We denote the input sequence of texts to be $S\{s_1,s_2,\ldots,s_{\vert S\vert}\}$, and we want to predict which category $y_i(i \in [1,2,\ldots,N])$ the sequence $S$ belongs to. At the same time, the model can also produce generative fine-grained explanations $e_c$ for $y_i$. 

\section{Generative Explanation Framework}
\begin{figure}[th]
	\centering
	\includegraphics[width=0.45\textwidth]{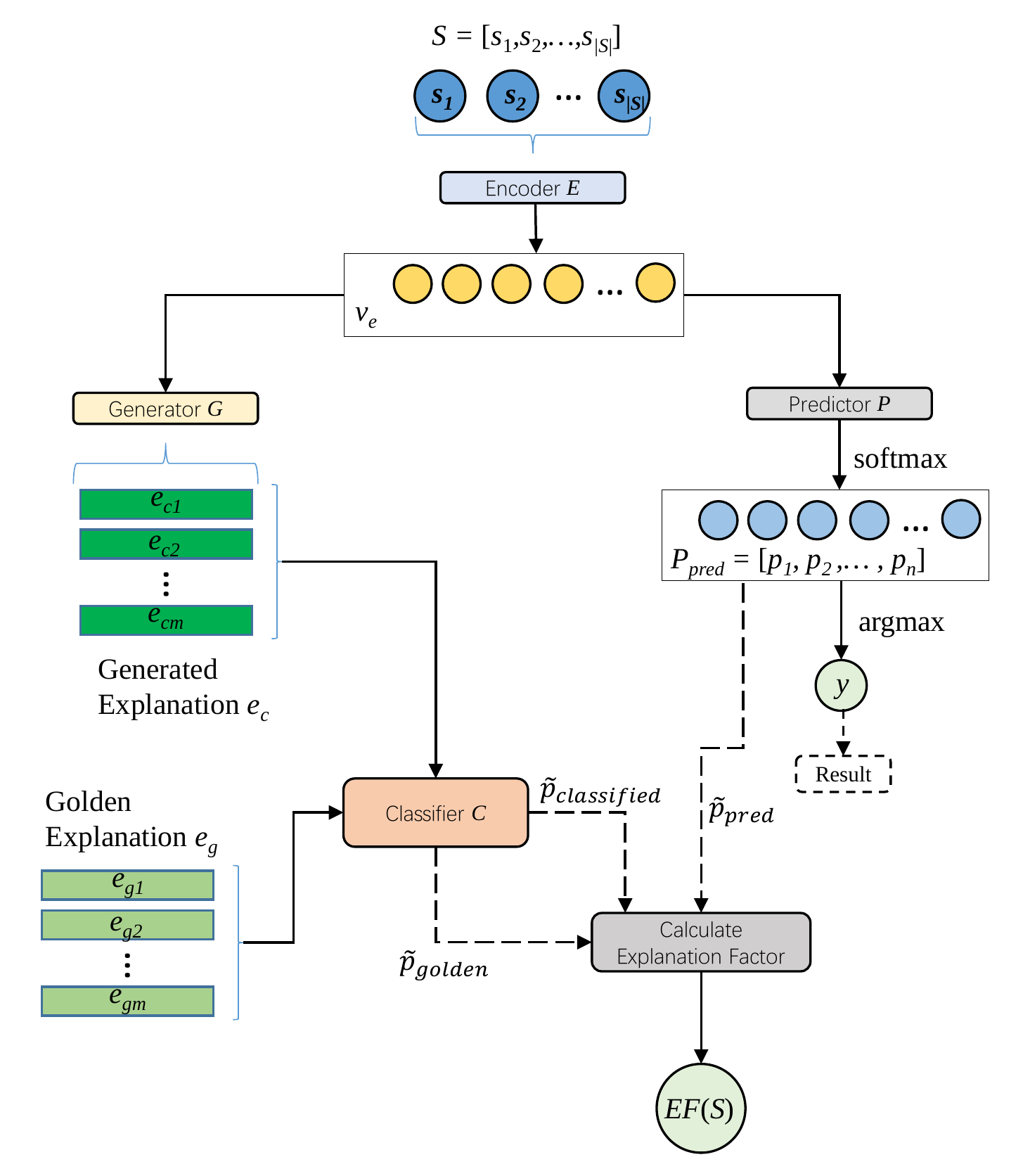}
	\caption{The architecture of the Generative Explanation Framework. $E$ encodes $S$ into a representation vector $v_e$. $P$ gives the probability distribution $P_{pred}$ for categories. We extract the ground-truth probability $\tilde{p}_{pred}$ from $P_{pred}$. Generator $G$ takes $v_e$ as input and generates explanations $e_c$. Classifier $C$ and Predictor $P$ both predict classes $y$. $C$ will predict a probability distribution $P_{classified}$ when taking $e_c$ as input, and predict $P_{golden}$ when taking $e_g$ as input, and then output the ground-truth probability $\tilde{p}_{classified}$ and $\tilde{p}_{golden}$. The explanation factor $EF(S)$ is calculated through $\tilde{p}_{pred}$, $\tilde{p}_{classified}$ and $\tilde{p}_{golden}$.}
	\label{archi_gen}
\end{figure}

In this part, we introduce our proposed Generative Explanation Framework (GEF). Figure \ref{archi_gen} illustrates the architecture of our model.

\subsection{Base Classifier and Generator}
A common way to do text classification tasks is using an Encoder-Predictor architecture \cite{zhang2015character,lai2015recurrent}.
As shown in Figure \ref{archi_gen}, a text encoder $E$ takes the input text sequence $S$, and encodes $S$ into a representation vector $v_e$. A category predictor $P$ then gets $v_e$ as input and outputs the category $y_i$ and its corresponding probability distribution $P_{pred}$.

As mentioned above, a desirable model should not only predict the overall results $y_i$, but also provide generative explanations to illustrate why it makes such predictions. A simple way to generate explanations is to feed $v_e$ to an explanation generator $G$ to generate fine-grained explanations $e_c$. This procedure is formulated as:
\begin{gather}
v_e = Encoder([s_1,s_2,\cdots,s_{\vert S\vert}])\\
P_{pred} = Predictor(v_e)\\
y = \mathop{\arg\max}_{i}(P_{pred,i})\\
e_c = f_G(W_G\cdot v_e + b_G)
\end{gather}
where $Encoder$ maps the input sequence $[s_1,s_2,\cdots,s_{\vert S\vert}]$ into the representation vector $v_e$; the $Predictor$ takes the $v_e$ as input and outputs the probability distribution over classification categories by using the $softmax$.

During the training process, the overall loss $\mathcal{L}$ is composed of two parts, i.e., the classification loss $\mathcal{L}_{p}$ and explanation generation loss $\mathcal{L}_{e}$:
\begin{gather}
\mathcal{L}(e_g,S,\theta) = \mathcal{L}_p + \mathcal{L}_e
\label{overall_loss}
\end{gather}
where $\theta$ represents all the parameters.

\subsection{Explanation Factor}
The simple supervised way to generate explanations, as demonstrated in the previous subsection, is quite straightforward. However, there is a significant shortcoming of this generating process: it fails to build strong connections between the generative explanations and the predicted overall results. In other words, the generative explanations seem to be independent of the predicted overall results. Therefore, in order to generate more reasonable explanations for the results, we propose to use an explanation factor to help build stronger connections between the explanations and predictions.

As we have demonstrated in the introduction section, fine-grained information will sometimes reflect the overall results more intuitively than the original input text sequence. For example, given a review sentence, ``The product is good to use'', we may not be sure if the product should be rated as 5 stars or 4 stars. However, if we see that the attributes of the given product are all rated as 5 stars, we may be more convinced that the overall rating for the product should be 5 stars.

So in the first place, we pre-train a classifier $C$, which also learns to predict the category $y$ by directly taking the explanations as input. More specifically, the goal of $C$ is to imitate human beings' behavior, which means that $C$ should predict the overall results more accurately than the base model that takes the original text as the input. We prove this assumption in the experiments section.


We then use the pre-trained classifier $C$ to help provide a strong guidance for the text encoder $E$, making it capable of generating a more informative representation vector $v_e$. During the training process, we first get the generative explanations $e_c$ by utilizing the explanation generator $G$. We then feed this generative explanations $e_c$ to the classifier $C$ to get the probability distribution of the predicted results $P_{classified}$. Meanwhile, we can also get the golden probability distribution $P_{gold}$ by feeding the golden explanations $e_g$ to $C$. The process can be formulated as:
\begin{gather}
P_{classified} = softmax(f_C(W_C\cdot e_c + b_C))\\
P_{gold} = softmax(f_C(W_C\cdot e_g + b_C))
\end{gather}

In order to measure the distance among predicted results, generated explanations and golden generations, we extract the ground-truth probability $\tilde{p}_{classified}$, $\tilde{p}_{pred}$, $\tilde{p}_{gold}$ from $P_{classified}$, $P_{pred}$, $P_{gold}$ respectively. They will be used to measure the discrepancy between the predicted result and ground-truth result in minimum risk training. 

We define our explanation factor $EF(S)$ as:
\begin{equation}
\begin{split}
EF(S) = \vert \tilde{p}_{classified} &- \tilde{p}_{gold} \vert + \\
&\vert \tilde{p}_{classified} - \tilde{p}_{pred}\vert
\end{split}
\end{equation}
There are two components in this formula.
\begin{itemize}
    \item The first part $\vert \tilde{p}_{classified} - \tilde{p}_{gold} \vert$ represents the distance between the generated explanations $e_c$ and the golden explanations $e_g$. Since we pre-train $C$ using golden explanations, we hold the view that if similar explanations are fed to $C$, similar predictions should be generated. For instance, if we feed a golden explanation ``Great performance'' to the classifier $C$ and it tells that this explanation means ``a good product'', then we feed another explanation ``Excellent performance'' to $C$, it should also tell that the explanation means ``a good product''. For this task, we hope that $e_c$ can express the same or similar meaning as $e_g$.
    \item The second part $\vert \tilde{p}_{classified} - \tilde{p}_{pred}\vert$ represents the relevance between the generated explanations $e_c$ and the original texts $S$. The generated explanations should be able to interpret the overall result. For example, if the base model predicts $S$ to be ``a good product'', but the classifier tends to classify $e_c$ to be the explanations for ``a bad product'', then $e_c$ cannot properly explain the reason why the base model gives such predictions.
\end{itemize}

\subsection{Minimum Risk Training}
In order to remove the disconnection between fine-grained information and input text, we use Minimum risk training (MRT) to optimize our models, which aims to minimize the expected loss, i.e., risk over the training data \cite{ayana2016neural}.
Given a sequence $S$ and golden explanations $e_g$, we define $\mathcal{Y}(e_g,S,\theta)$ as the set of predicted overall results with parameter $\theta$. We define $\Delta(y, \tilde{y})$ as the semantic distance between predicted overall results $y$ and ground-truth $\tilde{y}$. Then, the objective function is defined as:
\begin{gather}
\mathcal{L}_{MRT}(e_g,S,\theta) = \sum_{(e_g, S)\in D} \mathbb{E}_{\mathcal{Y}(e_g,S,\theta)}\Delta(y,\tilde{y})
\end{gather}
where $D$ presents the whole training dataset.

In our experiment, $\mathbb{E}_{\mathcal{Y}(e_g,S,\theta)}$ is the expectation over the set $\mathcal{Y}(e_g,S,\theta)$, which is the overall loss in Equation \ref{overall_loss}. And we define Explanation Factor $EF(S)$ as the semantic distance of input texts, generated explanations and golden explanations. Therefore, the objective function of MRT can be further formalized as:
\begin{gather}
\mathcal{L}_{MRT}(e_g,S,\theta) = \sum_{(e_g, S)\in D}\mathcal{L}(e_g,S,\theta)EF(S)
\end{gather}

MRT exploits $EF(S)$ to measure the loss, which learns to optimize GEF with respect to the specific evaluation metrics of the task. Though $\mathcal{L}_{MRT}$ can be 0 or close to 0 when $\tilde{p}_{classified}$, $\tilde{p}_{pred}$ and $\tilde{p}_{gold}$ are close, this cannot guarantee that generated explanations are close to the golden explanations. In order to avoid the total degradation of loss, we define our final loss function as the sum of MRT loss and explanation generation loss:
\begin{gather}
\mathcal{L}_{final} = \sum_{(e_g, S)\in D}\mathcal{L} + \mathcal{L}_{MRT}
\end{gather}

We try different weighting scheme for the overall loss, and get best performance with $1:$1.



\subsection{Application Case}
Generally, the fine-grained explanations are in different forms for a real-world dataset, which means that $e_c$ can be in the form of texts or in the form of numerical scores. We apply GEF to both forms of explanations using different base models.
\subsubsection{Case 1: Text Explanations}
 To test the performance of GEF on generating text explanations, we apply GEF to Conditional Variational Autoencoder (CVAE) \cite{sohn2015learning}. We here utilize CVAE because we want to generate explanations conditioned on different emotions (positive, negative and neural) and CVAE is found to be capable of generating emotional texts and capturing greater diversity than traditional SEQ2SEQ models.

\begin{figure}[t]
	\centering
	\includegraphics[width=0.45\textwidth]{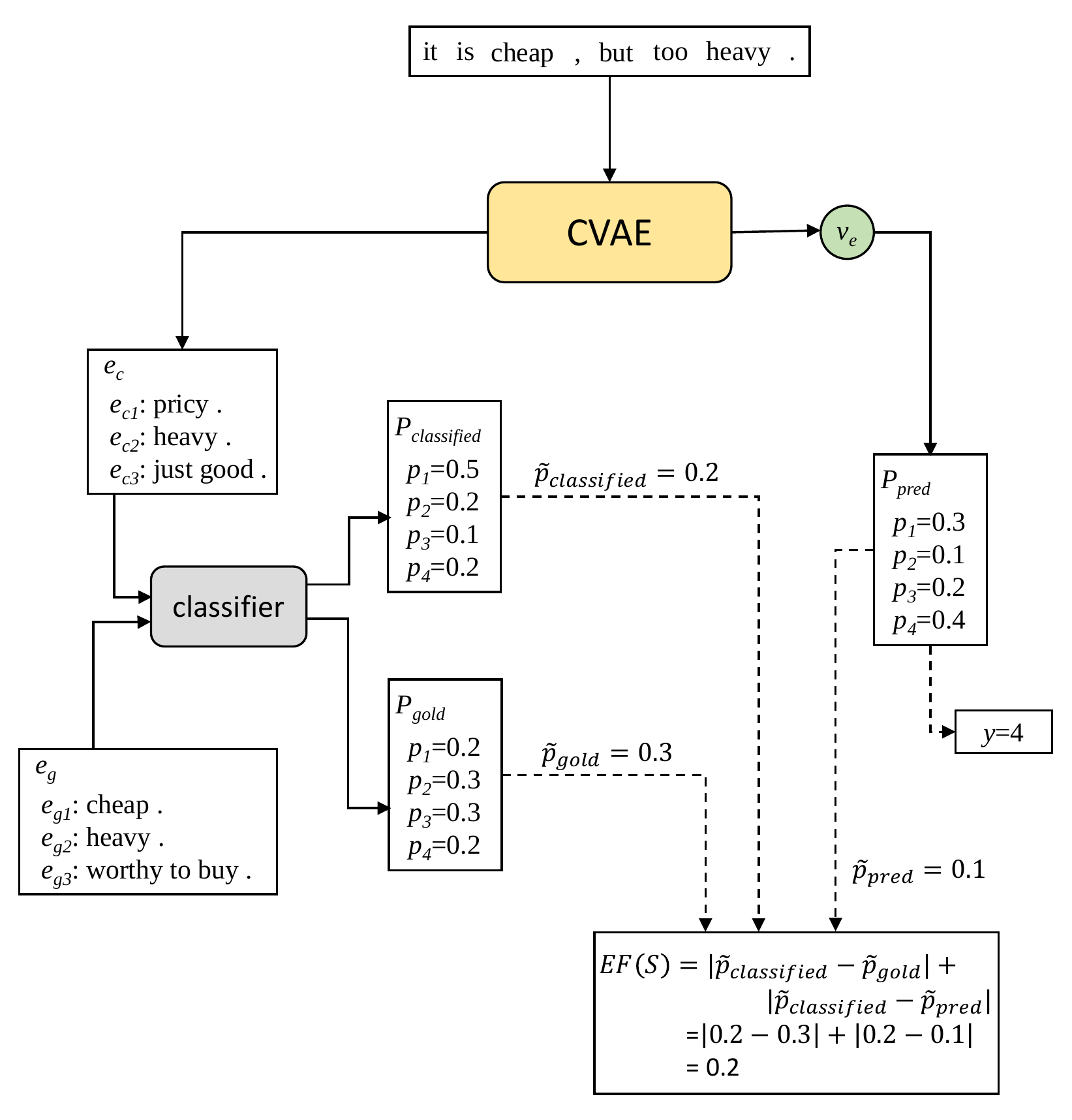}
	\caption{Structure of CVAE+GEF. There are totally 4 categories for the classification, and the ground-truth category is 2 in this example. We assume that the pre-trained classifier is a "perfect" classifier that will correctly predict the final label to be 2 when taking $e_g$ as input. So we wish the classifier can also predict the final result as label 2 when taking $e_c$ as input. This is why we focus on $\tilde{p}_{classified}$ and $\tilde{p}_{gold}$.}
	\label{cvae_gef}
\end{figure}

We give an example of the structure of CVAE+GEF in Figure \ref{cvae_gef}. For space consideration, we leave out the detailed structure of CVAE, and will elaborate it in the supplementary materials. In this architecture, golden explanations $e_g$ and generated explanations $e_c$ are both composed of three text comments: positive comments, negative comments, and neutral comments, which are fine-grained explanations for the final overall rating. The classifier is a skip-connected model of bidirectional GRU-RNN layers \cite{felbo2017using}. It takes three kinds of comments as inputs, and outputs the probability distribution over the predicted classifications.

\subsubsection{Case 2: Numerical Explanations}
Another frequently employed form of the fine-grained explanations for the overall results is numerical scores. For example, when a user wants to rate a product, s/he may first rate some attributes of the product, like the packaging, price, etc. After rating all the attributes, s/he will give an overall rating for the product. So we can say that the rating for the attributes can somewhat explain why the user gives the overall rating.
LSTM and CNN are shown to achieve great performance in text classification tasks \cite{tang-qin-liu:2015:EMNLP}, so we use LSTM and CNN models as the encoder $E$ respectively. The numerical explanations are also regarded as a classification problem in this example.

\section{Dataset}
We conduct experiments on two datasets where we use texts and numerical ratings to represent fine-grained information respectively. The first one is crawled from a website called PCMag, and the other one is the Skytrax User Reviews Dataset.
Note that all the texts in the two datasets are preprocessed by the Stanford Tokenizer\footnote{\url{https://nlp.stanford.edu/software/tokenizer.html}} \cite{manning2014stanford}.

\begin{table*}
	\begin{center}\small
		\begin{tabular}{lccccccccc}
			\Xhline{1pt}
			Overall Score & 1.0 & 1.5 & 2.0 & 2.5 & 3.0 & 3.5 & 4.0 & 4.5 & 5.0 \\
			\hline
			Number & 21 & 60 & 283 & 809 & 2399 & 3981 & 4838 & 1179 & 78 \\
			\Xhline{1pt}
		\end{tabular}
		\caption{Distribution of examples by each overall rating score in PCMag Review Dataset.}
		\label{pcmag_statistic}
	\end{center}
\end{table*}

\begin{table*}
	\begin{center}\small
		\begin{tabular}{lcccccccccc}
			\Xhline{1pt}
			Overall Score & 1 & 2 & 3 & 4 & 5 & 6 & 7 & 8 & 9 & 10 \\
			\hline
			Number & 4073 & 2190 & 1724 & 1186 & 1821 & 1302 & 2387 & 3874 & 4008 & 4530 \\
			\Xhline{1pt}
		\end{tabular}
		\caption{Distribution of examples by each overall rating score in Skytrax User Reviews Dataset.}
		\label{flight_statistic}
	\end{center}
\end{table*}

\subsection{PCMag Review Dataset}
This dataset is crawled from the website PCMag. It is a website providing reviews for electronic products, like laptops, smartphones, cameras and so on. Each item in the dataset consists of three parts: a long review text, three short comments, and an overall rating score for the product. Three short comments are summaries of the long review respectively from positive, negative, neutral perspectives. An overall rating score is a number ranging from 0 to 5, and the possible values that the score could be are \{1.0, 1.5, 2.0, ..., 5.0\}.

Since long text generation is not what we focus on, the items where review text contains more than 70 sentences or comments contain greater than 75 tokens are filtered. We randomly split the dataset into 10919/1373/1356 pairs for train/dev/test set. The distribution of the overall rating scores within this corpus is shown in Table \ref{pcmag_statistic}.

\subsection{Skytrax User Reviews Dataset}
We incorporate an airline review dataset scraped from Skytrax’s Web portal. Each item in this dataset consists of three parts: i.e., a review text, five sub-field scores and an overall rating score. The five sub-field scores respectively stand for the user's ratings for seat comfortability, cabin stuff, food, in-flight environment, and ticket value, and each score is an integer between 0 and 5. The overall score is an integer between 1 and 10.

Similar to the PCMag Review Dataset, we filter out the items where the review contains more than 300 tokens. Then we randomly split the dataset into 21676/2710/2709 pairs for train/dev/test set. The distribution of the overall rating scores within this corpus is shown in Table \ref{flight_statistic}.

\section{Experiments and Analysis}

\subsection{Experimental Settings}

As the goal of this study is to propose an explanation framework, in order to test the effectiveness of proposed GEF, we use the same experimental settings on the base model and on the base model+GEF.
We use GloVe \cite{pennington2014glove} word embedding for PCMag dataset and minimize the objective function using Adam \cite{kingma2014adam}. The hyperparameter settings for both datasets are listed in Table \ref{experiment_setting}. Meanwhile, since the generation loss is larger than classification loss for text explanations, we stop updating the predictor after classification loss reaches a certain threshold (adjusted based on dev set) to avoid overfitting.
\begin{table}[t]
	\begin{center}\small
		\begin{tabular}{lccc}
			\Xhline{1pt}
			& Embedding & hidden & batch size\\
			\hline
			PCMag & GloVe, 100 & 128  & 32\\
			Skytrax & random, 100 & 256 & 64\\
			\Xhline{1pt}
		\end{tabular}
		\caption{\label{experiment_setting}  Experimental settings for our experiments. Note that for CNN, we additionally set filter number to be $256$ and filter sizes to be $[3,4,5,6]$.}
	\end{center}
\end{table}


\subsection{Experimental Results}

\subsubsection{Results of Text Explanations}
We use BLEU \cite{papineni2002bleu} scores to evaluation the quality of generated text explanations. Table \ref{bleu_res} shows the comparison results of explanations generated by CVAE and CVAE+GEF.
\begin{table}[htpb]
	\begin{center}\small
		\begin{tabular}{llccccc}
			\Xhline{1pt}
			& & {\scriptsize BLEU-1} & {\scriptsize BLEU-2} & {\scriptsize BLEU-3} & {\scriptsize BLEU-4}\\
			\Xhline{1pt}
			\multirow{2}{*}{Pos.} & {\scriptsize CVAE} & 36.1 & 13.5 & 3.7  & 2.2 \\
			& {\scriptsize CVAE+GEF} & {\bf 40.1} & {\bf 15.6} & {\bf 4.5} & {\bf 2.6} \\
			\hline
			\multirow{2}{*}{Neg.} & {\scriptsize CVAE} & 33.3 & 14.1 & 3.1  &  2.2 \\
			& {\scriptsize CVAE+GEF} & {\bf 35.9} & {\bf 16.0} & {\bf 4.0} & {\bf 2.9} \\
			\hline
			\multirow{2}{*}{Neu.} & {\scriptsize CVAE} & 30.0 & 8.8 & 2.0 &  1.2 \\
			& {\scriptsize CVAE+GEF} & {\bf 33.2} & {\bf 10.2} & {\bf 2.5} & {\bf 1.5} \\
			\Xhline{1pt}
		\end{tabular}
		\caption{\label{bleu_res}  BLEU scores for generated explanations. Pos., Neg., Neu. respectively stand for positive, negative and neural explanations. The low BLEU-3 and BLEU-4 scores are because the target explanations contain many domain-specific words with low frequency, which makes it hard for the model to generate accurate explanations.}
	\end{center}
\end{table}


There are considerable improvements on the BLEU scores of explanations generated by CVAE+GEF over the explanations generated by CVAE, which demonstrates that the explanations generated by CVAE+GEF are of higher quality. CVAE+GEF can generate explanations that are closer to the overall results, thus can better illustrate why our model makes such a decision. 

In our opinion, the generated fine-grained explanations should provide the extra guidance to the classification task, so we also compare the performance of classification on CVAE and CVAE+GEF.  
We use top-1 accuracy and top-3 accuracy as the evaluation metrics for the performance of classification. In Table \ref{pcmag_classify}, we compare the results of CVAE+GEF with CVAE in both test and dev set. As shown in the table, CVAE+GEF has better classification results than CVAE, which indicates that the fine-grained information can really help enhance the overall classification results. 

\begin{table}[t]
	\begin{center}\small
		\begin{tabular}{lcc}
			\Xhline{1pt}
			& Acc\% (Dev) & Acc\% (Test)\\
			\Xhline{1pt}
			CVAE & 42.07 & 42.58  \\
			CVAE+GEF & {\bf 44.04} & {\bf 43.67} \\
			\hline
			\emph{Oracle} & 46.43 & 46.73 \\
			\Xhline{1pt}
		\end{tabular}
		\caption{\label{pcmag_classify}  Classification accuracy on PCMag Review Dataset. \emph{Oracle} means if we feed ground-truth text explanations to the Classifier $C$, the accuracy $C$ can achieve to do classification. \emph{Oracle} confirms our assumption that explanations can do better in classification than the original text.}
	\end{center}
\end{table}

As aforementioned, we have an assumption that if we use fine-grained explanations for classification, we shall get better results than using the original input texts. Therefore, we list the performance of the classifier $C$ in Table \ref{pcmag_classify} to make the comparison. Experiments show that $C$ has better performance than both CVAE and CVAE+GEF, which proves our assumption to be reasonable.

\subsubsection{Results of Numerical Explanations}
In the Skytrax User Reviews Dataset, the overall ratings are integers between 1 to 10, and the five sub-field ratings are integers between 0 and 5. All of them can be treated as classification problems, so we use accuracy to evaluate the performance.

The accuracy of predicting the sub-field ratings can indicate the quality of generated numerical explanations. In order to prove that GEF can help generate better explanations, we show the accuracy of the sub-field rating classification in Table \ref{field_res}. The 5 ratings evaluate the seat comfortability, cabin stuff, food, in-flight environment, and ticket value, respectively.
\begin{table}[t]
	\begin{center}\small
		\begin{tabular}{lccccc}
			\Xhline{1pt}
			& s\% & c\% & f\% & i\% & t\% \\
			\Xhline{1pt}
			LSTM & 46.59 & 52.27 & 43.74 & 41.82 & 45.04 \\
			LSTM+GEF & {\bf 49.13} & {\bf 53.16} & {\bf 46.29} & {\bf 42.34} & {\bf 48.25}\\
			\hline
			CNN & 46.22 & 51.83 & 44.59 & 43.34 & 46.88 \\
			CNN+GEF & {\bf 49.80} & {\bf 52.49} & {\bf 48.03} & {\bf 44.67} & {\bf 48.76} \\
			\Xhline{1pt}
		\end{tabular}
		\caption{\label{field_res}  Accuracy of sub-field numerical explanations on Skytrax User Reviews Dataset. s, c, f, t, v stand for seat comfortability, cabin stuff, food, in-flight environment and ticket value, respectively.}
	\end{center}
\end{table}
As we can see from the results in Table \ref{field_res}, the accuracy for 5 sub-field ratings all get enhanced comparing with the baseline. Therefore, we can tell that GEF can improve the quality of generated numerical explanations. 

Then we compare the result for classification in Table \ref{flight_classify}. As the table shows, the accuracy or top-3 accuracy both get improved when the models are combined with GEF.

Moreover, the performances of the classifier are better than LSTM (+GEF) and CNN (+GEF), which further confirms our assumption that the classifier $C$ can imitate the conceptual habits of human beings. Leveraging the explanations can provide guidance for the model when doing final results prediction.

\begin{table}[htpb]
	\begin{center}\small
		\begin{tabular}{lcccc}
			\Xhline{1pt}
			& Acc\% & Top-3 Acc\% \\
			\Xhline{1pt}
			LSTM & 38.06 & 76.89 \\
			LSTM+GEF & {\bf 39.20} & {\bf 77.96} \\
			\hline
			CNN & 37.06 & 76.85 \\
			CNN+GEF & {\bf 39.02} & {\bf 79.07} \\
			\hline
			\emph{Oracle} & 45.00 & 83.13 \\
			\Xhline{1pt}
		\end{tabular}
		\caption{\label{flight_classify} Classification accuracy on Skytrax User Reviews Dataset. \emph{Oracle} means if we feed ground-truth numerical explanation to the Classifier $C$, the accuracy $C$ can achieve to do classification.}
	\end{center}
\end{table}

\subsection{Human Evaluation}
In order to prove our model-agnostic framework can make the basic model generate explanations more closely aligned with the classification results, we employ crowdsourced judges to evaluate a random sample of $100$ items in the form of text, each being assigned to $5$ judges on the Amazon Mechanical Turk. All the items are correctly classified both using the basic model and using GEF, so that we can clearly compare the explainability of these generated text explanations. We report the results in Table \ref{human_evaluation}, and we can see that over half of the judges think that our GEF can generate explanations more related to the classification results. In particular, for $57.62\%$ of the tested items, our GEF can generate better or equal explanations comparing with the basic model. 

In addition, we show some the examples of text explanations generated by CVAE+GEF in Table \ref{text_example}. We can see that our model can accurately capture some key points in the golden explanations. And it can learn to generate grammatical comments that are logically reasonable. All these illustrate the efficient of our method. We will demonstrate more of our results in the supplementary materials.

\begin{table}[t]
	\begin{center}\small
		\begin{tabular}{lcccc}
			\Xhline{1pt}
			& Win\% & Lose\% & Tie\% \\
			\Xhline{1pt}
			CVAE+GEF & 51.37 & 42.38 & 6.25 \\
			\Xhline{1pt}
		\end{tabular}
		\caption{\label{human_evaluation} Results of human evaluation. Tests are conducted between the text explanations generated by basic CVAE and CVAE+GEF.}
	\end{center}
\end{table}

\subsection{Error and Analysis}
\newcommand{\tabincell}[2]{\begin{tabular}{@{}#1@{}}#2\end{tabular}}
\begin{table*}
	\begin{center}\small
		\begin{tabular}{p{2cm} p{13cm}}
			\Xhline{1pt}
			\bf Product and Overall Rating & \bf Explanations \\
			\Xhline{1pt}
			\multirow{5}{*}{Monitor, 3.0} & \tabincell{p{14cm}}{{\bf Positive Generated:} \textcolor{red}{very affordable. \underline{unique} and ergonomic \underline{design}. \underline{good port} selection.} \\
				{\bf Positive Golden:} \textcolor{blue}{\underline{unique design}. dual hdmi \underline{ports}. \underline{good} color quality. energy efficient.}} \\
			& \\
			& \tabincell{p{13cm}}{{\bf Negative Generated:} \textcolor{red}{relatively faint on some \underline{features}. relatively high contrast ratio. no auto port.} \\ 
				{\bf Negative Golden: } \textcolor{blue}{expensive. weak light \underline{grayscale performance}. features are scarce.}} \\
			& \\
			& \tabincell{p{13cm}}{{\bf Neutral Generated:} \textcolor{red}{\underline{the samsung series is a} unique touch-screen \underline{monitor} featuring a unique design and a nice capacitive picture, but its \underline{color} and \underline{grayscale performance} could be better.} \\
				{\bf Neutral Golden:} \textcolor{blue}{\underline{the samsung series is a} stylish 27-inch \underline{monitor} offering good color reproduction and sharp image quality. however, it 's more expensive than most tn monitors and has a limited \underline{feature} set.}} \\
			\Xhline{1pt}
		\end{tabular}
		\caption{\label{text_example}  Examples of our generated explanations. Some key points are underlined.}
	\end{center}
\end{table*}


We focus on the deficiency of generation for text explanation in this part. 

First of all, as we can see from Table \ref{text_example}, the generated text explanation tend to be shorter than golden explanations. It is because longer explanations tend to bring more loss, so GEF tends to leave out the words that are of less informative, like function words, conjunctions, etc. In order to solve this problem, we may consider adding length reward/penalty by reinforcement learning to control the length of generated texts.

Second, there are $\langle$UNK$\rangle$s in the generated explanations. Since we are generating abstractive comments for product reviews, there may exist some domain-specific words. The frequency of these special words is low, so it is relatively hard for GEF to learn to embed and generated these words. A substituted way is that we can use copy-mechanism \cite{gu2016incorporating} to generate these domain-specific words.

\section{Related Work}
Our work is closely aligned with Explainable Artificial Intelligence \cite{gunning2017explainable}, which is claimed to be essential if users are to understand, and effectively manage this incoming generation of artificially intelligent partners. In artificial intelligence, providing an explanation of individual decisions has attracted attention in recent years. The traditional way of explaining the results is to build connections between the input and output, and figure out how much each dimension or element contributes to the final output. Some previous works explain the result in two ways: evaluating the sensitivity of output if input changes and analyzing the result from a mathematical perspective by redistributing the prediction function backward \cite{samek2017explainable}. There are some works connecting the result with the classification model. \newcite{ribeiro2016should} selects a set of representative instances with explanations via submodular optimization. Although the method is promising and mathematically reasonable, they cannot generate explanations in natural forms. They focus on how to interpret the result.

Some of the previous works have similar motivations as our work. \newcite{lei2016rationalizing} rationalize neural prediction by extracting the phrases from the input texts as explanations. They conduct their work in an extractive way, and focus on rationalizing the predictions. However, our work aims not only to predict the results but also to generate abstractive explanations, and our framework can generate explanations both in the forms of texts and numerical scores. \newcite{P18-1175} proposes to use a classifier with natural language explanations that are annotated by human beings to do the classification. Our work is different from theirs in that we use the natural attributes as the explanations which are more frequent in reality. \newcite{NIPS2018_8163} proposes e-SNLI\footnote{The dataset is not publicly available now. We would like to conduct further experiments on this dataset when it is released.} by extending SNLI dataset with text explanations. And their simple but effective model proves the feasibility of generating text explanations for neural classification models.


\section{Conclusion}
In this paper, we investigate the possibility of using fine-grained information to help explain the decision made by our classification model. More specifically, we design a Generative Explanation Framework (GEF) that can be adapted to different models. Minimum risk training method is applied to our proposed framework. Experiments demonstrate that after combining with GEF, the performance of the base model can be enhanced. Meanwhile, the quality of explanations generated by our model is also improved, which demonstrates that GEF is capable of generating more reasonable explanations for the decision.

Since our proposed framework is model-agnostic, we can combine it with other natural processing tasks, e.g. summarization, extraction, which we leave to our future work.

\bibliography{acl2019}
\bibliographystyle{acl_natbib}

\clearpage

\section*{Supplemental Material}
\label{sec:supplemental}
\subsection*{Structure of CVAE}
\begin{figure}[th]
	\centering

	\includegraphics[width=0.45\textwidth]{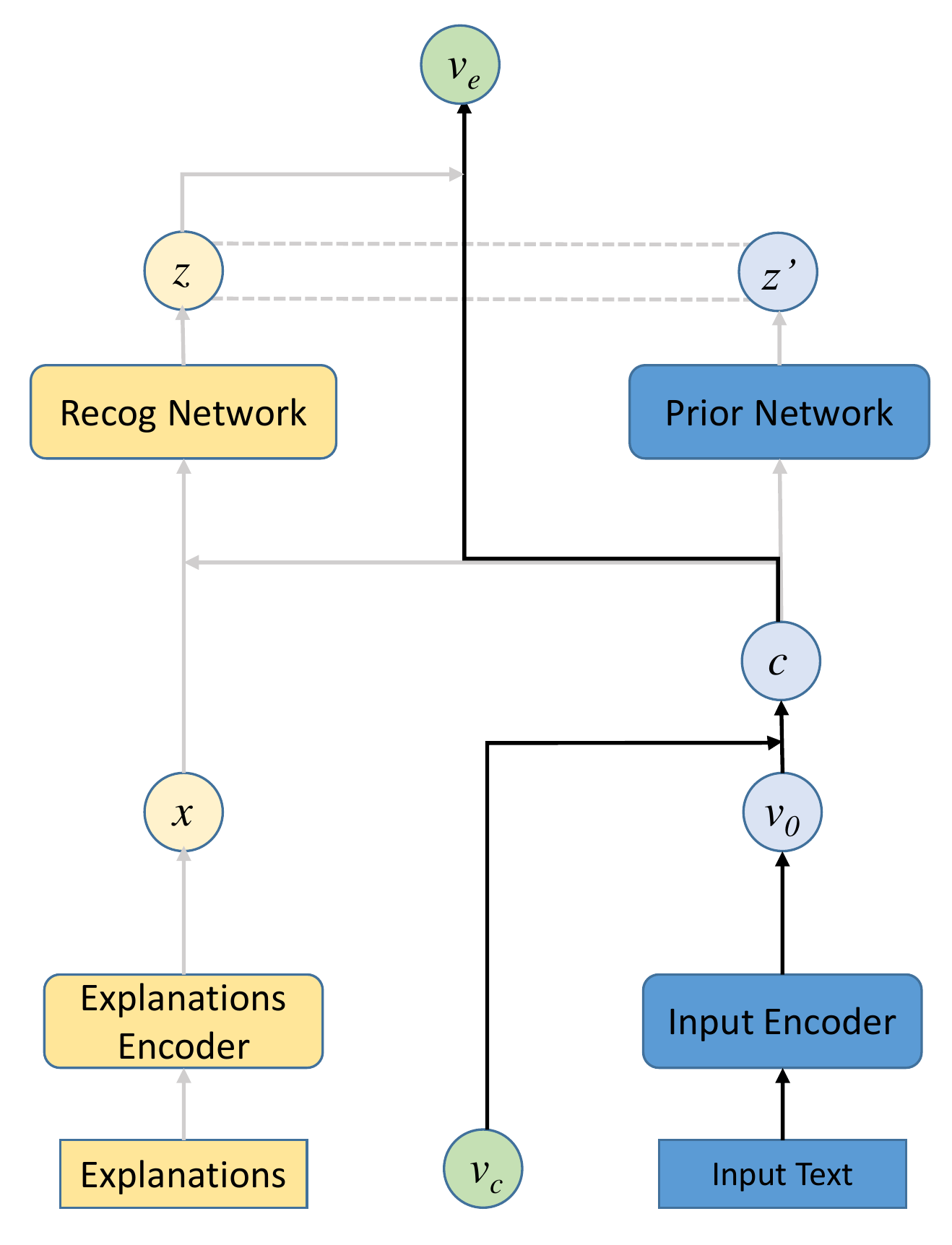}
	\caption{The structure of CVAE.  The Input Encoder encodes the input text in $v_0$, and $v_c$ is the control signal that determines the kind of fine-grained information (positive, negative and neutral). $v_e$ is the initial input for the decoder. The Explanations Encoder encodes the short comment in $x$. Recognition Network takes $x$ as input and produces the latent variable $z$. In our experiment, the Recognition Network and the Prior Network are both MLPs, and we use bidirectional GRU as the Explanations Encoder and Input Encoder.}
	\label{cvae_strucure}
\end{figure}

By extending the SEQ2SEQ structure, we can easily get a Conditional Variational Antoencoder (CVAE) \cite{sohn2015learning,acl2018zhou}. Figure \ref{cvae_strucure} shows the structure of the model.

\appendix

To train CVAE, we need to maximize a variational lower bound on the conditional likelihood of $x$ given $c$, where $x$ and $c$ are both random variables. In our experiment,$c=[v_c;v_0]$, and $x$ is the text explanations we want to generate. This can be rewritten as:
\begin{gather}
    p(x\vert c) = \int p(x\vert z, c)p(z\vert c)dz
\end{gather}
$z$ is the latent variable. The decoder is used to approximate $p(x\vert z, c)$, denoted as $p_D(x\vert z, c)$, and Prior Network is used to approximate $p(z\vert c)$, denoted as $p_P(z\vert c)$. In order to approximate the true posterior $p(z\vert x, c)$, we introduce Recognition Network $q_R(z\vert x, c)$. According to \newcite{sohn2015learning}, we can have the lower bound of $\log p(x\vert c)$ as:
\begin{multline}
    -\mathcal{L}(x, c; \theta) = KL(q_R(z\vert x, c)\vert\vert p_P(z\vert c)) \\
    -\mathbb{E}_{q_R(z\vert x, c)}(\log p_D(x\vert z, c))
\end{multline}
$\theta$ is the parameters in the network. Notice that during training, $z$ is used to train $z^{\prime}$ and passed to the decoder, but during testing, the ground truth explanations are absent and $z^{\prime}$ is passed to the decoder.

\subsection*{Output Sample}
In this part, we provide some samples from our experiment.
\subsubsection*{Numerical Explanation Cases}
We provide some numerical explanation cases in Table \ref{number_example}.
\begin{table}[t]
	\begin{center}\small
		\begin{tabular}{lcccccc}
			\Xhline{1pt}
			Overall & & s & c & f & i & t \\
			\Xhline{1pt}
			\multirow{2}{*}{9.0} & {\bf pred} & 4.0 & 5.0 & 5.0 & 4.0 & 5.0 \\
			& {\bf gold} & 4.0 & 5.0 & 5.0 & 4.0 & 4.0\\
			\hline
			\multirow{2}{*}{6.0} & {\bf pred} & 3.0 & 5.0 & 3.0 & 3.0 & 4.0 \\
			& {\bf gold} & 4.0 & 5.0 & 3.0 & 3.0 & 4.0\\
			\hline
			\multirow{2}{*}{2.0} & {\bf pred} & 2.0 & 1.0 & 2.0 & 2.0 & 2.0 \\
			& {\bf gold} & 2.0 & 2.0 & 1.0 & 2.0 & 2.0\\
			\Xhline{1pt}
		\end{tabular}
		\caption{\label{number_example} Examples from the results on Skytrax User Reviews Dataset. s, c, f, i, t stand for seat comfortability, cabin stuff, food, in-flight environment and ticket value, respectively.}
	\end{center}
\end{table}

\subsubsection*{Text Explanation Cases}
We provide some text explanation cases in Table \ref{text_example}.
\begin{table*}
	\begin{center}\small
		\begin{tabular}{p{2cm} p{13cm}}
			\Xhline{1pt}
			\bf Product and Overall Rating & \bf Explanations \\
			\Xhline{1pt}
			\multirow{5}{*}{Television, 4.0} & \tabincell{p{12cm}}{{\bf Positive Generated:} \textcolor{red}{Good contrast. Good black levels. Affordable.}\\
				{\bf Positive Golden:} \textcolor{blue}{Gorgeous 4k picture. Good color accuracy. Solid value for a large uhd screen.}} \\
			& \\
			& \tabincell{p{12cm}}{{\bf Negative Generated:} \textcolor{red}{Mediocre black levels. Poor shadow detail. Poor off-angle viewing.}\\ 
				{\bf Negative Golden:} \textcolor{blue}{Mediocre black levels. Poor input lag. Colors run slightly cool. Disappointing online features. Poor off-angle viewing.}} \\
			& \\
			& \tabincell{p{12cm}}{{\bf Neutral Generated:} \textcolor{red}{A solid, gorgeous 4k screen that offers a sharp 4k picture, but it's missing some features for the competition.} \\
				{\bf Neutral Golden:} \textcolor{blue}{A solid 4k television line, but you can get an excellent 1080p screen with more features and better performance for much less.}} \\
			\hline
			\multirow{5}{*}{Flash Drive, 3.0} & \tabincell{p{12cm}}{{\bf Positive Generated:} \textcolor{red}{Simple, functional design. Handy features.} \\
				{\bf Positive Golden:} \textcolor{blue}{Charming design. Reasonably priced. Capless design.}} \\
			& \\
			& \tabincell{p{12cm}}{{\bf Negative Generated:} \textcolor{red}{All-plastic construction. No usb or color protection.} \\ 
				{\bf Negative Golden: } \textcolor{blue}{All-plastic construction. On the slow side. Crowds neighboring ports. flash drives geared toward younger children don't have some sort of password protection.}} \\
			& \\
			& \tabincell{p{12cm}}{{\bf Neutral Generated:} \textcolor{red}{The tween-friendly $\langle$UNK$\rangle$ colorbytes are clearly designed and offers a comprehensive usb 3.0, but it's not as good as the competition.} \\
				{\bf Neutral Golden:} \textcolor{blue}{The kid-friendly dane-elec sharebytes value pack drives aren't the quickest or most rugged flash drives out there, but they manage to strike the balance between toy and technology. Careful parents would be better off giving their children flash drives with some sort of password protection.}} \\
			\hline
			\multirow{5}{*}{TV, 4.0} & \tabincell{p{12cm}}{{\bf Positive Generated:} \textcolor{red}{excellent picture. attractive glass-backed screen. hdr10 and dolby vision.} \\
				{\bf Positive Golden:} \textcolor{blue}{excellent picture with wide color gamut. stylish glass-backed screen. hdr10 and dolby vision. two remotes.}} \\
			& \\
			& \tabincell{p{12cm}}{{\bf Negative Generated:} \textcolor{red}{very expensive.} \\ 
				{\bf Negative Golden: } \textcolor{blue}{very expensive.}} \\
			& \\
			& \tabincell{p{12cm}}{{\bf Neutral Generated:} \textcolor{red}{lg's new oledg7p series is a stylish, attractive, and attractive hdtv line that's a bit more but not much more attractive.} \\
				{\bf Neutral Golden:} \textcolor{blue}{lg's signature oledg7p series is every bit as attractive and capable as last year's excellent oledg6p series, but the company has a new flagship oled that's only slightly more expensive but a lot more impressive.}} \\
			\hline
			\multirow{5}{*}{Gaming, 4.0} & \tabincell{p{12cm}}{{\bf Positive Generated:} \textcolor{red}{best-looking mainline pokemon game for the nintendo 3ds and feel. date, breathing, and dlc.} \\
				{\bf Positive Golden:} \textcolor{blue}{best-looking mainline pokemon game to date. alola trials mix up and vary progression over the gym-and-badge system, breathing new life into the game for longtime fans. ride pagers improve overworld navigation.}} \\
			& \\
			& \tabincell{p{12cm}}{{\bf Negative Generated:} \textcolor{red}{starts out very slow.} \\ 
				{\bf Negative Golden: } \textcolor{blue}{starts out very slow.}} \\
			& \\
			& \tabincell{p{12cm}}{{\bf Neutral Generated:} \textcolor{red}{the newest pokemon generation of sun/moon for the nintendo 3ds, making the feeling of the nintendo 3ds and remixes enough ideas to new life over making any wild, polarizing changes to the formula.} \\
				{\bf Neutral Golden:} \textcolor{blue}{the newest pokemon generation, sun/moon for the nintendo 3ds, tweaks and polishes the series' core concepts and remixes enough ideas to feel fresh without making any wild , polarizing changes to the formula.}} \\
			\hline
			\multirow{5}{*}{Desktop, 3.5} & \tabincell{p{12cm}}{{\bf Positive Generated:} \textcolor{red}{adjustable bulb. attractive design. energy efficient.} \\
				{\bf Positive Golden:} \textcolor{blue}{compact all in one. \$500 price point. lenovo utilities. dynamic brightness system and eye distance system. no bloatware.}} \\
			& \\
			& \tabincell{p{12cm}}{{\bf Negative Generated:} \textcolor{red}{limited stand. no keyboard or micro between mac.} \\ 
				{\bf Negative Golden: } \textcolor{blue}{low power on benchmark tests. no usb 3.0. no hdmi. no video in or out. only 60-day mcafee anti-virus. camera is `` always on. ''.}} \\
			& \\
			& \tabincell{p{12cm}}{{\bf Neutral Generated:} \textcolor{red}{the lenovo thinkcentre edge is a good choice in the attractive design, and a few attractive colors in the price. it has a little bit of the best.} \\
				{\bf Neutral Golden:} \textcolor{blue}{the lenovo c325 is a good choice for those looking to spend only about \$500 for a fully featured desktop pc. it's bigger than a laptop, and has the power to serve your web surfing and basic pc needs.}} \\
			\Xhline{1pt}
		\end{tabular}
		\caption{\label{text_example}  Text examples from our generated explanations. $\langle$UNK$\rangle$ stands for ``unknown word''.}
	\end{center}
\end{table*}

\end{document}